\documentclass[11pt]{article}

\usepackage[preprint]{acl}

\usepackage{times}
\usepackage{latexsym}

\usepackage[T1]{fontenc}

\usepackage[utf8]{inputenc}

\usepackage{microtype}

\usepackage{inconsolata}

\usepackage{graphicx}

\usepackage{amsmath,amssymb,amsfonts}
\usepackage{algorithmic}
\usepackage{graphicx}
\usepackage{textcomp}
\usepackage{subcaption}
\usepackage{multirow}
\usepackage{caption} 
\usepackage{booktabs}
\usepackage[table]{xcolor} 
\usepackage{xcolor}
\definecolor{mygreen}{RGB}{0,128,0}
\definecolor{myred}{RGB}{255,0,0} 
\definecolor{lightgray}{gray}{0.92}
\definecolor{lightyellow}{rgb}{1,1,0.85}

\title{Beyond Single-Dimensional Compression: The Compound Sparsity Frontier of Large Language Models}

\author{
 \textbf{Chao Han},
 \textbf{Haozhe Hu},
 \textbf{Xiaoyu Shen}\thanks{Corresponding Author}
\\
\\
 Ningbo Institute of Digital Twin, Eastern Institute of Technology, Ningbo
\\
 \small{
   \textbf{Correspondence: Xiaoyu Shen} \href{mailto:email@domain}{xyshen@eitech.edu.cn}
 }
}

\begin{document}
\maketitle
\begin{abstract}
Large language models (LLMs) are often compressed through static parameter pruning or dynamic token-level computation, yet aggressive sparsification can trigger rapid performance degradation beyond an essential sparsity boundary. This work asks \emph{whether combining these two mechanisms can delay such degradation by distributing the compression burden}. We study a minimalist compound sparsity framework that first applies low-rank approximation and channel pruning to obtain a statically compressed backbone, and then introduces lightweight routers for per-token dynamic layer skipping. This design enables independent control of parameter sparsity and token-level computation sparsity. Experiments across language understanding and modeling benchmarks show that compound sparsity consistently outperforms single-mechanism compression under the same total sparsity, delaying the decay point on understanding tasks and preserving stronger modeling performance. Further analysis reveals cross-dimensional interference between parameter pruning and token skipping, and shows that near-balanced allocation is most effective under a fixed sparsity budget. These results demonstrate that compound compression provides a practical way to improve LLM compression, while revealing a broader cross-dimensional sparsity boundary that ultimately limits further compression. Code will be available at https://github.com/EIT-NLP/LLM-Pruning.
\end{abstract}

\section{Introduction}
Scaling large language models (LLMs) has led to remarkable progress in language understanding and generation~\cite{brown2020language,su2022welm,su2024unraveling,ahn2024large,lin2025explore,zhang2025llm}. However, their high inference costs make deployment difficult in resource-constrained environments. Model compression has therefore become a central direction for reducing memory usage and computation while preserving model capability~\cite{han2016deep,fan2025visipruner}.

A key challenge in LLM compression is the existence of \emph{essential sparsity}. Compression can usually remove part of the model or computation with limited degradation, but after a critical point, further sparsification causes rapid performance decay~\cite{NEURIPS2023_7a69ab48,jaiswal2024compressing,ding2026llms,wu2026hidrop}.

This sparsity boundary is usually studied through individual compression methods. In LLM compression, two representative directions are static parameter pruning and dynamic token-level computation. Static parameter pruning reduces the structural cost of the model by removing or factorizing parameters before inference~\cite{frantar2023sparsegpt,ainslie2024slicegpt, huang2025determining}. Dynamic token-level computation methods instead reduce inference cost during inference by allocating computation adaptively to different tokens or inputs~\cite{raposo2024mixture, jiang2024d, han2025informed,he2026skipopu}. Although these methods exploit different sources of redundancy, they are typically pushed and evaluated separately, so the resulting sparsity limit is often treated as a property of a single compression mechanism.

This single-mechanism view leaves an important question open: \emph{if one compression mechanism eventually reaches its sparsity boundary, can combining it with another mechanism delay the onset of degradation?} The intuition is simple. Redundancy in LLMs appears across multiple levels, including parameters, layers, and token-level computation~\cite{fulazyllm,dumitru2024change,liang2025entropy,han2026unirank}, so a target compression budget does not have to be imposed entirely on one source of redundancy. Instead, part of the budget can be assigned to static parameter pruning to reduce the cost of the backbone, while another part can be assigned to dynamic token skipping to reduce input-dependent computation. By distributing compression pressure, compound compression may achieve the same total sparsity while keeping each individual mechanism in a less aggressive regime.

In this work, we use the combination of static parameter pruning and dynamic token skipping as a representative and controllable case study for this question. Concretely, we first apply LoRAP~\cite{li2024lorap} and channel pruning to obtain a statically compressed backbone, and then add lightweight routers for per-token dynamic layer skipping~\cite{zhao2025skipgpt}. This design allows us to independently control parameter sparsity and token-level computation sparsity, making it possible to analyze how different compression allocations affect the sparsity-performance trade-off.
\begin{figure*}[htbp]
	\centering 
	\includegraphics[width=0.95\textwidth]{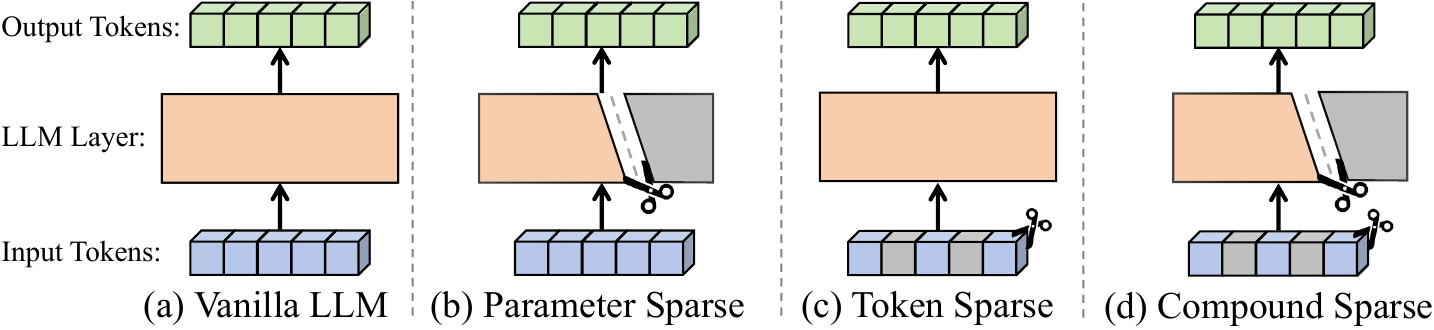}  
	\caption{Overview of LLMs compression paradigms, compound sparse distributes sparsity burdens into orthogonal parameter and token space.}  
	\label{fig:framework}
\end{figure*}

Our experiments show that compound sparsity consistently achieves better compression-performance trade-offs than single-mechanism compression under the same total sparsity. On understanding tasks, it delays the sharp decay point from around 20\% to around 30\% sparsity. On language modeling, it also retains higher performance, e.g., about 70\% relative performance at 30\% sparsity, compared with about 60\% for token skipping and 56\% for parameter pruning. Further analysis reveals cross-dimensional interference: a mildly pruned backbone becomes more sensitive to token sparsity. Under a fixed 50\% total sparsity budget, near-balanced allocation achieves the best performance. These findings support the \emph{Two Dimensions, One Wall} principle: compound compression delays performance decay, but a broader cross-dimensional sparsity boundary remains.

Our contributions are threefold. (1) We study essential sparsity from a compound compression perspective, asking whether combining static parameter pruning and dynamic token skipping can delay performance decay. (2) We introduce a minimalist and controllable framework based on LoRAP, channel pruning, and per-token dynamic layer skipping, enabling independent control of parameter and token sparsity. (3) We show that compound sparsity improves the compression-performance trade-off and delays decay compared with single-mechanism compression, while still converging to a cross-dimensional sparsity boundary.

\section{Methodology}
Our goal is to build a standardized, controllable paradigm for probing compound sparsity, to explore the interplay and fundamental limits between parameter sparsity and token-level dynamic computation sparsity. We adopt two established methods: LoRAP~\cite{li2024lorap} for static parameter pruning and SkipGPT~\cite{zhao2025skipgpt} for token-wise dynamic layer skipping. Using off-the-shelf techniques rules out biases from bespoke designs, ensuring our observations purely reflect the effects of cross-dimensional sparsity allocation.

\subsection{Compound Compression Pipeline}
We first formulate the forward computation paradigm within a standard transformer layer of pre-trained LLMs. Denote the original forward mapping of an uncompressed transformer layer (as Figure \ref{fig:framework}(a) shown) as:
\begin{equation}
	\boldsymbol{y} = \mathcal{F}(\boldsymbol{x})
\end{equation}
where $\boldsymbol{x} \in \mathbb{R}^{N\times d}$ denotes the input token embedding sequence with sequence length $N$ and hidden dimension $d$, and $\mathcal{F}(\cdot)$ represents the full computational operator of the intact transformer layer.

\paragraph{Step 1: Static Parameter Space Compression.}
We first conduct structured parameter pruning upon the original layer operator $\mathcal{F}(\cdot)$. Let $s_p \in [0,1]$ denote the \textit{parameter sparsity ratio}, which quantifies the proportion of pruned redundant parameters within the layer. After parameter pruning, the original full-layer operator is converted into a sparse parameter backbone operator $\mathcal{F}_p(\cdot)$. Correspondingly, the overall computational overhead of this layer is reduced to $(1-s_p)$ of the original magnitude. The forward propagation after parameter compression (see Figure \ref{fig:framework}(b)) is formulated as:
\begin{equation}
	\boldsymbol{y}_p = \mathcal{F}_p(\boldsymbol{x})
\end{equation}

In our implementation, we adopt LoRAP~\cite{li2024lorap} for parameter pruning, which applies low-rank decomposition to attention modules and channel pruning to MLP modules.

\paragraph{Step 2: Dynamic Token-level Computational Compression.}
Built upon the pruned sparse backbone, we further integrate lightweight inference-adaptive routing mechanism to perform token-wise computational skipping. We define $s_t \in [0,1]$ as the \textit{token computational sparsity ratio}, indicating the expected proportion of tokens that are exempted from full-layer forward computation.

Following SkipGPT~\cite{zhao2025skipgpt}, a lightweight post-trained routing module is deployed within each transformer layer to produce binary routing indicators for every individual token. The entire input token sequence $\boldsymbol{x}$ is divided into two disjoint subsets (corresponding to blue and gray cubes in Figure \ref{fig:framework}(c)) according to routing decisions: Active token subset $\boldsymbol{x}_\mathcal{A}$ which execute forward computation and Inactive token subset $\boldsymbol{x}_\mathcal{I}$ which direct delivered to subsequent layer.

The final hybrid forward computation of the compound sparse layer can be formally written in piece-wise form:
\begin{equation}
	\boldsymbol{y}_{\text{t}} =
	\begin{cases}
		\mathcal{F}_p(\boldsymbol{x}), & \boldsymbol{x} \in \boldsymbol{x}_\mathcal{A} \\
		\boldsymbol{x}, & \boldsymbol{x} \in \boldsymbol{x}_\mathcal{I}
	\end{cases}
\end{equation}
This hierarchical two-stage compression paradigm strictly separates parameter structural sparsity and token adaptive computational sparsity, and realizes decoupled control over sparsity intensity in two orthogonal optimization dimensions.

\subsection{Definition of Compound Sparsity}
In this work, we uniformly define sparsity from the perspective of \emph{effective inference computation reduction ratio}, which maintains consistent measurement criteria across different compression dimensions \cite{raposo2024mixture, jiang2024d} and facilitates fair comparison.

After cascaded parameter pruning and dynamic token skipping, the retained computational volume of the target layer is quantified as the product of reserved computation ratio in two dimensions:
\[
\text{Retained Computation Ratio} = (1-s_p) \cdot (1-s_t)
\]
Accordingly, we define the \emph{global compound sparsity degree} $S_{\text{comp}}$ as the total proportion of eliminated computation resources:
\begin{equation}
	S_{\text{comp}} = 1 - (1-s_p)(1-s_t)
\end{equation}

Under fixed target sparsity $S_{\text{target}}$, we can adjust the combination of $(s_p, s_t)$ to realize diversified sparsity allocation strategies: allocating more compression budget to parameter dimension, placing more sparsity burden on token dynamic dimension, or adopting balanced dual-dimensional allocation. Such flexible allocation strategy constitutes the core experimental basis for us to explore whether shifting compression pressure can evade single-dimensional sparsity collapse limitation.

\subsection{Balanced Sparsity Allocation}
We adopt uniform sparsity assignment for balanced compression budget distribution. Given parameter sparsity $s_p$ and token sparsity $s_t$, the overall compound sparsity is defined as $S_{\text{comp}}=1-(1-s_p)(1-s_t)$. For a targeted global sparsity level, we evenly split compression intensity across two dimensions by setting $s_p=s_t=X$. Solving $(1-X)^2 = 1-S_{\text{comp}}$ yields the unified sparsity ratio. Mathematically, equal allocation minimizes the maximum sparsity imposed on either dimension, which serves as our fundamental allocation principle in experiments. Section \ref{sec:sparsity_allocation} shows such paradigm also works well in practical. 

\section{Experiments}
\subsection{Setup}
We use Llama3.1-8b as the base model and Redpajama as training data.
We evaluate model performance on both understanding and modeling tasks.
\textit{Understanding}:Accuracy on BoolQ \cite{clark2019boolq}, PIQA \cite{bisk2020piqa}, HellaSwag \cite{zellers2019hellaswag}, Winogrande \cite{sakaguchi2021winogrande}, ARC-E/ARC-C \cite{clark2018think}, and OBQA \cite{mihaylov-etal-2018-suit}. \textit{Modeling}: Perplexity (PPL) on WikiText-2 \cite{Merity2016PointerSM}. All tasks are evaluated by lm-eval 0.4.9 \cite{eval-harness}. We report relative performance (e.g. compressed accuracy/original accuracy) to intuitively quantify the preservation of model capabilities. 
\subsection{Main Results}
\begin{figure}[htbp]   
	\centering
	
	\begin{subfigure}{0.48\textwidth}
		\centering
		\includegraphics[width=\linewidth]{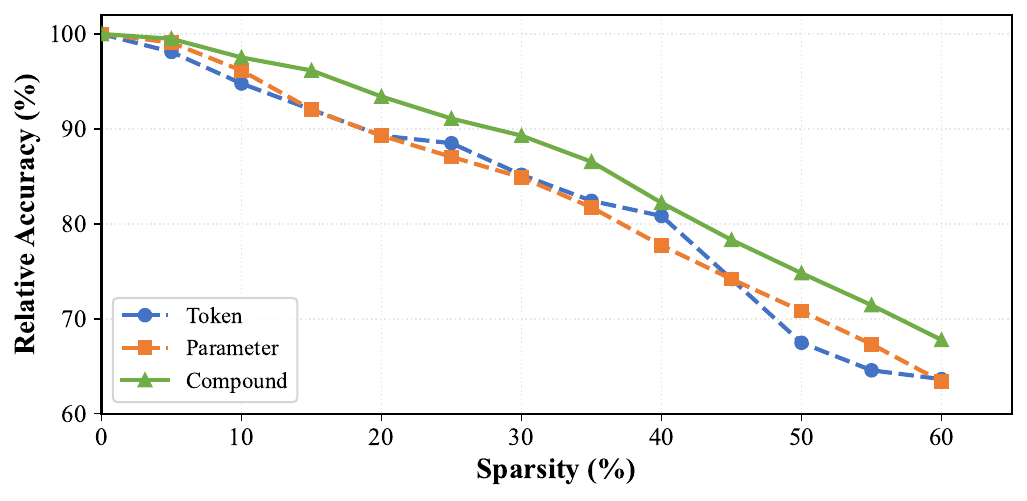}
		\caption{Performance degradation curve of understanding tasks.}
		\label{fig:sub1}
	\end{subfigure}
	\begin{subfigure}{0.48\textwidth}
		\centering
		\includegraphics[width=\linewidth]{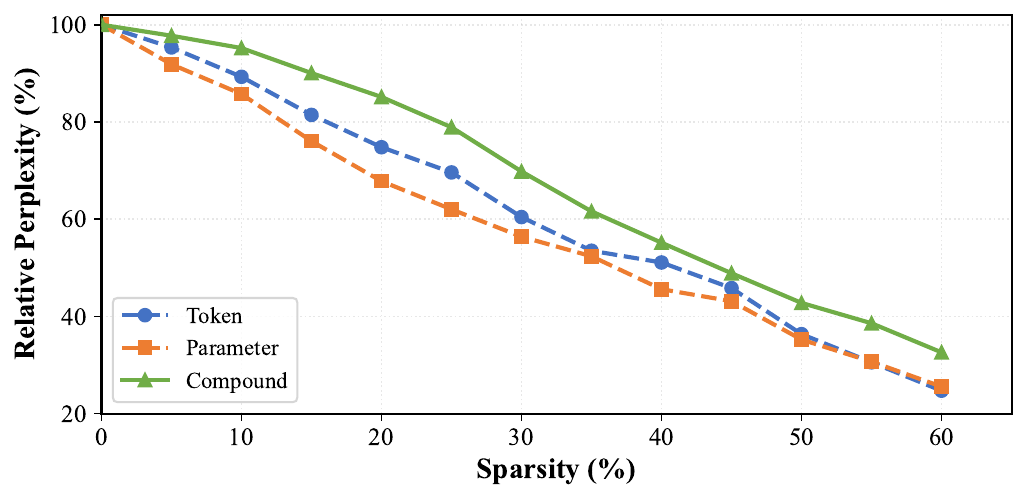}
		\caption{Performance degradation curve of modeling tasks.}
		\label{fig:sub2}
	\end{subfigure}
	
	\caption{Sparsity-performance landscape of different sparse paradigms.}
	\label{fig:main_res}
\end{figure}

\paragraph{Modeling is more sensitive.}
First, language modeling suffers severer performance degradation than language understanding. The difference lies in task nature: multiple-choice understanding tasks are robust to compression noise due to simple linguistic patterns, while modeling tasks demand precise context prediction and are highly sensitive to sparsity-induced knowledge loss.

\paragraph{Decay is delayed.}
Second, the essential sparsity phenomenon is consistently observed in compound compression, with distinct degradation patterns on understanding tasks. Taking 90\% relative performance as the critical boundary, understanding tasks only lose 10\% performance within 0-30\% sparsity but suffer a sharp 25\% performance drop at 30–60\% sparsity. Notably, the compound sparsity paradigm delays the essential sparsity collapse threshold from 20\% (single-dimensional compression) to a higher sparsity range. Differently, language modeling only exhibits early-stage essential sparsity within 0–10\% sparsity, followed by approximate linear degradation, forming an inverse scaling curve.

\paragraph{Compound sparsity performs best.}
Third, dual-dimensional sparsity allocation yields persistent performance gains under identical total compression sparsity. Compared with single-dimensional compression, compound sparsity strategy distributes compression budgets across parameter and token dimensions, fully excavating and utilizing the heterogeneous redundant information of LLMs parameter and token space. This orthogonal redundancy utilization avoids over-compression on a single dimension, mitigates premature performance collapse, and thus achieves better trade-offs between compression ratio and model performance.

\subsection{Further Studies}
\paragraph{Cross-dimensional interference.}

\begin{table}[ht]
	\centering
	\small
	\caption{Performance of Different Methods under Varying Sparsity Ratios}
	\label{tab:sparsity_performance}
	\resizebox{0.5\textwidth}{!}{
		\begin{tabular}{lcccccc}
			\toprule
			\multirow{2}{*}{Base} & \multicolumn{6}{c}{Sparsity (\%)} \\
			\cmidrule{2-7}
			& 0 & 10 & 20 & 30 & 40 & 50 \\
			\midrule
			Original & 72.81 & 69.42 & 65.17 & 62.42 & 58.84 & 49.11 \\
			Compressed & 69.49 & 66.71 & 61.36 & 55.92 & 49.15 & 41.23 \\
			Pres. (\%) & 95.44 & 96.09 & 94.15 & 89.54 & 83.53 & 83.95 \\
			\bottomrule
		\end{tabular}
	}
\end{table}

We further explore the interplay between parameter and token sparsity by comparing performance degradation curves of dynamic token skipping on the dense baseline and the model with 10\% parameter sparsity. While parameter pruning and token skipping independently reduce computational overhead, we observe evident cross-dimensional interference in practical performance. Even the 10\% parameter-sparse model, which retains 95\% performance of the original model, becomes more vulnerable to token compression. When token sparsity reaches 60\%, the compound sparse model only maintains 83\% baseline performance. Such cross-dimensional fragility reveals why a unified inherent limit exists for compound sparsity.

\paragraph{Allocation matters.}
\label{sec:sparsity_allocation}

\begin{table}[ht]
	\centering
	\caption{Sensitivity Analysis of Performance w.r.t. Parameter Sparsity}
	\label{tab:sparsity_correlation}
	\small 
	\begin{tabular}{lcccccc}
		\toprule
		$s_p$  & 0 & 10 & 20 & \textbf{30} & 40 & 50 \\
		\midrule
		$s_t$     & 50 & 44 & 38 & \textbf{28} & 17 & 0  \\
		\midrule
		Acc.    & 67.45 & 69.87 & 73.52 & \textbf{74.79} & 72.4 & 70.79 \\
		\bottomrule
	\end{tabular}
\end{table}

We next study how to allocate a fixed compression budget across the two dimensions. We analyze performance across cross-dimensional sparsity settings under 50\% total compound sparsity. As parameter sparsity increases from 0\% to 50\%, performance follows a convex curve: it improves with mild parameter sparsity but degrades drastically at high levels. Optimal results appear at a near-uniform allocation ($s_p\approx30\%, s_t\approx28.6\%$). This verifies our proposition that balanced cross-dimensional sparsity lowers single-dimensional compression intensity and achieves the best trade-off.


\section{Conclusion}
This work explores the fundamental limit of compound sparsity in LLMs compression using a concise, controllable framework that decouples static parameter pruning and dynamic token skipping. Our empirical evidence proves that sharing compression pressure across parameter and token dimensions cannot bypass the inherent sparsity limit, but merely defers performance degradation. A universal cross-dimensional sparsity wall remains the ultimate constraint. We also quantify sparsity-performance trends and identify distinct degradation behaviors across language tasks. Beyond revealing core compression principles, our findings provide principled references for cross-dimensional budget scheduling.

\section*{Limitations}
Our empirical analysis is conducted based on a limited set of LLM architectures and two representative compression paradigms, i.e., static parameter pruning and dynamic token skipping. Although we adopt classic and widely validated compression schemes to ensure credible observations, the current conclusions are not fully universal across arbitrary model architectures or diverse compression pipelines. The generalization of our discovered compound sparsity principles to other model families, scaling sizes, and advanced compression methods remains to be further verified. Future work can extend our probing framework to broader model settings and more compression variants to establish a more generalizable understanding of cross-dimensional sparsity limits.


\bibliography{2d1w}



\end{document}